\pdfoutput=1
\documentclass[twoside,11pt]{article}

\usepackage[nohyperref]{jmlr2e}

\usepackage{bbm}
\usepackage{dsfont}
\usepackage{algorithm}

\usepackage{mathtools}
\usepackage{multirow}
\usepackage{amsmath} 
\usepackage{booktabs}
\usepackage{tabularx}
\usepackage{makecell}
\usepackage{float}
\usepackage{hyperref}
\hypersetup{
    colorlinks=true,        
    linkcolor=blue,         
    citecolor=blue,         
    urlcolor=blue,          
    pdftitle={TNLearn: An Open Source Python Package for Task-based Neurons},
    pdfauthor={Meng Wang, Tieyun Li, Juntong Fan, Hanyu Pei, Jing-Xiao Liao, Yaodong Yang, Jianwei Ma, and Fenglei Fan},
}
\usepackage{doi}

\usepackage{natbib}
\usepackage{pythonhighlight}
\def\x{\mathbf{x}}
\def\X{\mathbf{X}}
\def\Y{\mathbf{Y}}

\def\w{\mathbf{w}}

\firstpageno{1}

\makeatletter
\def\@starteditor{}
\def\@endeditor{}
\makeatother

\begin{document}

\title{TNLearn: An Open Source Python Package for Task-based Neurons}

\author{\name Meng Wang$^{1*}$ \email wm\_ai@qq.com
       \AND
       \name Tieyun Li$^{2*}$ \email tieyun.li@stu.hit.edu.cn
       \AND
       \name Juntong Fan$^{3*}$ \email cuhkfanjuntong@gmail.com 
       \AND  
       \name Hanyu Pei$^3$ \email 1155183430@link.cuhk.edu.hk
              \AND
       \name Jing-Xiao Liao$^3$ \email jingxiao.liao@cityu.edu.hk 
               \AND 
       \name Yaodong Yang$^{4}$ \email Yaodong.yang@outlook.com    
       \AND 
        \name Jianwei Ma$^{2\dag}$ \email jma@hit.edu.cn 
       \AND  
       \name Fenglei Fan$^{3\dag}$ \email fenglfan@cityu.edu.hk
 \\             
       \addr $^1$Shenzhen H\&T Intelligent Control Co., Ltd., Shenzhen, China\\
       \addr $^2$School of Mathematics, 
       Harbin Institute of Technology, 
       Harbin, China \\
       \addr $^3$Department of Data Science, 
       The City University of Hong Kong, 
       Kowloon, Hong Kong  \\     \addr $^4$Institute of Artificial Intelligence, 
       Peking University, 
       Beijing, China \\
       \addr $^*$Meng Wang, Tieyun Li, and Juntong Fan contribute equally. $^\dag$Co-corresponding authors}

\editor{}

\maketitle

\begin{abstract}
The brain does not rely on a single type of neuron to perform all kinds of tasks; instead, it designs different neurons for different tasks. The concept of task-based neurons represents a paradigm shift compared to task-based architectures. It argues that solving a specific problem requires customized neurons, as task-based neurons capture useful prior knowledge from task-related data. To facilitate the use of task-based neurons in scientific research and industrial applications, we introduce \texttt{TNLearn}, an open-source Python package that provides automated construction of task-based neurons and networks, enabling smooth training of task-based networks. Comprehensive documentation, including technical exposition, API reference, and representative examples, is available online. 
\texttt{TNLearn} is open-sourced at \url{https://github.com/NewT123-WM/tnlearn} and has become a PyTorch ecosystem project.
\end{abstract}

\begin{keywords}
  Task-based neurons, NeuroAI, symbolic regression, Python package
\end{keywords}

\section{Introduction}

Over the past few years, many successful networks were due to the novel architecture designs \citep{srivastava2015highway, he2016deep, huang2017densely, fan2021sparse}. These networks almost exclusively use the same type of neurons that are based on the inner-product and the nonlinear activation function. This is in sharp contrast with our brain, which is made up of many functionally and morphologically various neurons \citep{peng2021morphological}. Moreover, in our brain, all kinds of intelligent behaviors are based on the coordination of different neurons. Therefore, can we introduce different types of neurons into artificial networks and examine what merits neuronal diversity can bring \citep{fan2025towards,fan2026no}? The core philosophy of introducing neuronal diversity is “dimension augmentation". An artificial network has two highly complementary dimensions: the neuron type and architecture. Designing well-performing neurons represents a paradigm shift relative to designing well-performing architectures. Moreover, the well-performing neurons can be integrated into the well-performing architecture to further boost a model's representation power.

Along this direction, there is a growing interest in introducing new neurons into deep learning \citep{chrysos2021deep, fan2018new, jiang2020nonlinear, mantini2021cqnn, goyal2020improved} such as quadratic neurons \citep{fan2025one, jiang2020nonlinear, mantini2021cqnn, goyal2020improved, liao2023attention}. \citet{xu2022quadralib} developed a package called \texttt{QuadraLib} for the optimization of quadratic neurons. Also, \citet{xu2026quadranet} proposed to use quadratic adaptation to train a high-order network. At the same time, high-order and quadratic neurons have been successfully applied in a wide spectrum of applications including medical imaging \citep{fan2019quadratic}, industrial informatics \citep{tang2024deep}, bearing fault diagnosis \citep{liao2025classifier}, clinical imaging diagnosis \citep{li2024application} and civil engineering \citep{nguyen2019deep}. 
This line of studies largely confirms the feasibility and potential of developing deep learning with new neurons.

Furthermore, we believe that there should be no one-size-fits-all neuron, \textit{i.e.}, the performance of one neuron cannot be the best for all tasks. Biological neuronal diversity emerges from the brain's necessity to tackle complex tasks. Rather than relying on a single neuron type for all tasks, the brain tailors neurons to specific tasks with neuronal differentiation. Therefore, the concept of task-based neurons differs from that of task-based architectures. The former asserts that customized neurons should be prototyped to address specific problems. Task-based neurons leverage valuable prior knowledge from task-related data, allowing networks composed of such neurons to integrate task-specific priors. Consequently, networks made of task-based neurons should outperform those built from generic neurons with the same structure. While the concepts of task-based neurons and networks have been discussed in previous works \citep{chrysos2021deep,chrysos2022augmenting, fan2026no}, limited effort has been devoted to systematically materialize these ideas, methodologies, and algorithms into user-friendly open-source software.

To tackle this challenge, we have developed 
\texttt{TNLearn}, an open-source Python library tailored for creating task-based neurons with different search methods and integrating them into networks. Serving as an initial implementation of task-based neurons, 
\texttt{TNLearn} is crafted to offer a user-friendly solution for beginners while also serving as a cutting-edge benchmark for seasoned researchers.
Primarily, \texttt{TNLearn} emphasizes user-friendliness, striking a harmonious balance between accessibility and flexibility. The architecture of 
\texttt{TNLearn} is modular, enabling users to selectively incorporate modules that align with their specific needs and workflows. As demonstrated in the code snippet below, building a task-based network in 
\texttt{TNLearn} is straightforward. With only a few lines of code, users can swiftly generate preliminary outcomes. This simplicity not only encourages broad adoption but also enriches users' machine learning capabilities, bringing this novel tool to neural network novices. Furthermore, 
\texttt{TNLearn} can function as a cutting-edge reference point, allowing users to benchmark their algorithmic designs effectively. We believe that 
\texttt{TNLearn}, as a part of the ecosystem, is instrumental to the research and translation of artificial neural networks.

\section{Design and Features of TNLearn}

Leveraging Python, a popular language in the machine learning community, 
\texttt{TNLearn} benefits from a rich ecosystem that provides many useful machine learning utilities and supports automatic differentiation. At a high level, 
\texttt{TNLearn} is designed to offer user-friendly strategies for easy instantiation and training with minimal code. Our implementations undergo rigorous automatic unit testing to ensure reliability. 
Figure~\ref{fig:tnlearn_arch} presents a high-level overview of 
\texttt{TNLearn}'s architecture. Now, we elaborate on each element:

\begin{figure}[H]
    \centering
    \includegraphics[width=\textwidth]{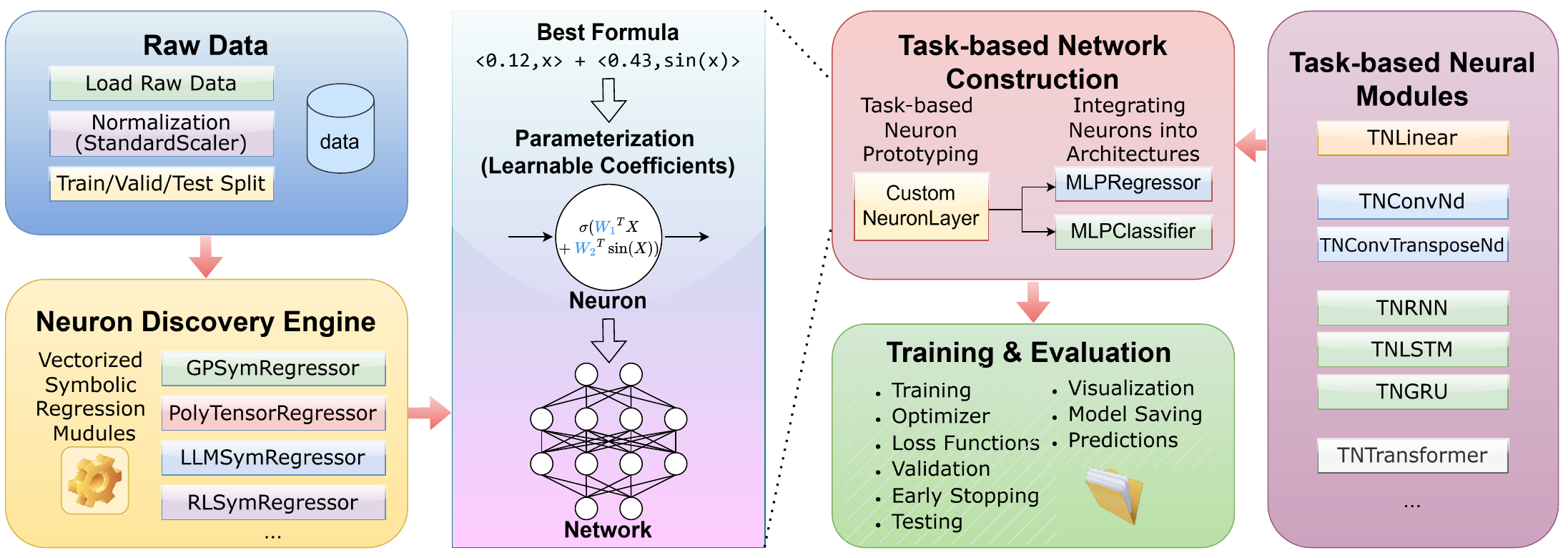}
    \vspace{-0.3cm}
    \caption{The high-level architecture of TNLearn}
\label{fig:tnlearn_arch}
\end{figure}

\subsection{Neuron Discovery Engine}
\noindent \textbf{Vectorized Symbolic Regression.} 
Symbolic regression (SR) is inspired by scientific discoveries in physics \citep{schmidt2009distilling, bartlett2023exhaustive}. Its goal is to identify optimal formulas that fit the input data by utilizing arbitrary compositions of primitive mathematical operations applied to the input variables, including arithmetic operators and logarithmic, trigonometric, and exponential functions. SR represents a formula as a tree structure and applies genetic programming based on this structure. The vectorized symbolic regression (VSR) forces each input variable to perform the same computation in order to learn a homogeneous formula for downstream parallel computation and avoid the curse of dimensionality. Here we slightly abuse the concept of homogeneity, meaning that all variables undergo identical operations, and the associated coefficients may differ. This module implements VSR from scratch, rather than being a direct import from the existing mature symbolic regression methods. Furthermore, how to find an effective vectorized formula is an optimization problem that cannot be solved by gradient-based methods, since gradients cannot be derived. We have implemented four methods to enrich the armory of our package: \texttt{GPSymRegressor}, \texttt{LLMSymRegressor}, \texttt{PolyTensorRegressor}, and \texttt{RLSymRegressor}. At present, there is no principle guiding which method is superior under which conditions. We encourage users to choose freely among them. In Appendix \ref{techniques}, we introduce all these neuron discovery methods in detail for readers' further reference. 

\begin{python}
from tnlearn import GPSymRegressor
from tnlearn import MLPRegressor
from sklearn.datasets import make_regression
from sklearn.model_selection import train_test_split
# Generate data.
X, y = make_regression(n_samples=200, random_state=1)
X_train, X_test, y_train, y_test = train_test_split(X, y, random_state=1)
# A symbolic regression algorithm is used to generate task-based neurons.
neuron = GPSymRegressor()
neuron.fit(X_train, y_train)
\end{python}

\texttt{GPSymRegressor} uses the genetic programming to generate random expressions, evaluate their fitness, and evolve expressions through mutation and crossover. It includes all necessary hyperparameters, such as the population size, tournament configuration, number of generations, and more, to assist users in optimizing the learning of formulas. In the example above, we use a \texttt{scikit-learn} built-in function \texttt{make\_regression} to generate synthetic data, and then apply \texttt{GPSymRegressor} to learn a homogeneous formula.

\texttt{LLMSymRegressor} employs a large language model to discover a formula. During training on massive data, large language models gain the so-called super-intuition for mathematical structures, enabling a direct ``guess'' at equation forms.
In \texttt{TNLearn}, this capability is integrated into the vectorized symbolic regression framework via \texttt{LLMSymRegressor}. Inspired by DrSR \citep{wang2025drsr}, \texttt{LLMSymRegressor} combines data-driven insights with the experience-oriented reflection, generates equation skeletons, and fits coefficients with optimizers like BFGS.

\texttt{RLSymRegressor} provides an efficient way to discover neuron formulas. It uses a policy network with risk-seeking REINFORCE \citep{williams1992simple} to search over the predefined candidate library, and fits the coefficients of the selected terms via Ridge regression. Users can control the search space via parameters such as the maximum power order \(K\), the maximum interaction order \(R\), and the maximum number of sampled terms \(m\) per episode, making it flexible for discovering formulas in various tasks while maintaining computational efficiency. Note that \texttt{RLSymRegressor} may be of independent interest to the field of symbolic regression, as it offers a reinforcement-learning perspective that recasts symbolic optimization as a sequential decision-making problem.

\texttt{PolyTensorRegressor} provides a concrete implementation within this framework by restricting the learned formula to a tensor-decomposed representation. It leverages tensor decomposition to efficiently discover symbolic formulas within the vectorized framework by assuming the maximum degree of the learned formula. Specifically, it decomposes the aggregation function into three complementary streams: a polynomial stream for independent feature transformations, a tensor-decomposed interaction stream that captures high-order feature dependencies using Canonical Polyadic (CP) decomposition with low rank \citep{guo2011tensor}, and a periodic stream for cyclic patterns. By jointly optimizing structural gates and projection matrices through gradient descent, this method provides a concise way to discover effective neurons end-to-end.

\subsection{Task-based Neuron Construction}
\label{sec:task_neuron_construction}

This module parameterizes the acquired formula, making the parameters learnable before the formula serves as the aggregation function of the neuron. The learned parameters can serve as a good initialization, or we can randomly initialize all neurons. Parameterization enables task-based neurons to adapt and interact within a network. Activation functions, such as the ReLU function \citep{nair2010rectified}, keep intact when connected to a network, as these functions have been widely tested and proven to perform well. This module also encourages users to customize the activation function.

\subsection{Task-based Network Construction}

\noindent \textbf{Integrating Task-based Neurons into an Architecture.}
After the task-based neurons are prototyped, we integrate these neurons into a specified network architecture for regression and classification tasks. This module follows the \texttt{scikit-learn} estimator API for data loading, loss definition, and evaluation, and relies on PyTorch for automatic differentiation. An example usage is as follows:

\begin{python}
# Build neural network using task-based neurons and train it.
clf = MLPRegressor(neurons=neuron.neuron)
clf.fit(X_train, y_train)
# Predict
clf.predict(X_test)
\end{python}

\noindent \textbf{Customizable Neuron Layers for Generic Architectures.}
In addition to the multilayer perceptron, \texttt{TNLearn} provides a family of modular layers that cover fully-connected, convolutional, recurrent, and Transformer architectures, with the neuron aggregation function governed by a user-defined symbolic expression.
These layers are constructed as immediate substitutes for the corresponding PyTorch modules, permitting their integration into the existing models without altering the surrounding structure. Numeric coefficients in the symbolic expression are ignored, and each layer learns its own weights.
This design equips every neuron with a tunable nonlinearity that can reflect prior knowledge of the problem domain or the statistical properties of the data.
For instance, a practitioner working on image data can replace a conventional convolution with a customized version by writing the following code:
\begin{python}
from tnlearn.modules import TNConv2d
conv = TNConv2d(3, 64, 3, symbolic_expression='x + sin(x)')
\end{python}
which introduces a sinusoidal inductive bias into the feature extraction process.
Similarly, recurrent cells and Transformer encoder layers accept the same symbolic expression, enabling coherent neuron specification across different network components.
By exposing the internal operation of the neuron, \texttt{TNLearn} facilitates systematic investigation into how alternative aggregation functions affect predictive accuracy and generalization capacity, a direction that remains largely uncharted with fixed activation functions.
The library offers thorough documentation and illustrative examples for these advanced layers, thereby supporting rapid prototyping and reproducible research.

\texttt{TNLearn} is accompanied by comprehensive documentation that covers technical expositions, task-based neuron guidance, and representative examples. The main resources are summarized in Table~\ref{tab:quickstart}.

\begin{table}[H]
\centering
\small
\begin{tabular}{ll}
\toprule
Resource & Details \\
\midrule
Installation & \texttt{pip install tnlearn} \\
Documentation & \url{https://tnlearn-documentation.readthedocs.io} \\
Source code & \url{https://github.com/NewT123-WM/tnlearn} \\
PyPI & \url{https://pypi.org/project/tnlearn} \\
\bottomrule
\end{tabular}
\caption{Quick start resources for TNLearn.}
\label{tab:quickstart}
\end{table}

\section{What Can You Do with TNLearn?}

\noindent \textbf{Model Design.} It can be used to solve real-world problems. Task-based neurons and generic neurons are two different approaches in the field of artificial neural networks. We summarize some potential advantages of task-based neurons over generic neurons, although these advantages of task-based neurons over generic neurons can depend on the specific task and the design of the neural network. i) \textit{Efficiency}: Task-based neurons are designed to perform specific tasks, which can make them more efficient in terms of computational resources and time. ii) \textit{Specialization}: Task-based neurons can be specialized for specific tasks because they incorporate more prior knowledge, which can lead to better performance in those tasks. iii) \textit{Interpretability (mild)}: Task-based neurons can be easier to interpret, as their function is directly related to the task they are designed for.

\noindent \textbf{Benchmark.} It can be used in neural network research. Task-based neurons can serve as a baseline method for comparison with other models. It serves as a reference point against which the performance of more complex or advanced models can be measured. This is particularly useful in machine learning and data science, where it helps to establish a standard for evaluating the effectiveness of different models.

\section{Conclusion and Outlook}

In summary, we present \texttt{TNLearn}, a user-friendly Python package for task-based neurons that aims to be easy to use, versatile for different data, and performant on different tasks. \texttt{TNLearn} can engage the interest of a wide audience, since \texttt{TNLearn} provides a basic machine learning tool instead of focusing on a specific application field. We plan to keep this package up-to-date by including more architectures such as Mamba \citep{gu2023mamba}, and more utilities. We will also keep this package actively maintained. Recently, EvoGP \citep{wu2026enabling} has demonstrated efficient GPU acceleration for tree-based GP through tensorized tree encoding and adaptive parallelism. We also plan to equip GPSymRegressor with GPU parallelization in future releases, enabling it to scale to larger datasets while preserving its lightweight and extensible design for task-based neuron discovery. We enthusiastically welcome contributions and suggestions from the community.   

\newpage
\appendix
\section{Project Structure and Installation}

\subsection{Project Structure}

The \texttt{TNLearn} codebase is organized into a modular directory hierarchy. Core functionality resides in the \texttt{tnlearn/} package, which provides the regressor implementations, custom operator logic, network layer construction, and custom module integration. The remaining directories are dedicated to testing, benchmarking, and example usage.

\begin{python}
tnlearn/
|-- tnlearn/                  # Main package
|   |-- __init__.py           # Export public classes: GPSymRegressor, MLPRegressor, MLPClassifier, LLMSymRegressor, RLSymRegressor, ...
|   |-- base.py               # BaseModel class
|   |-- base1.py              # Legacy base (maintained for compatibility)
|   |-- TN_base.py            # Task-based network base classes
|   |-- operator/             # Customized operators and simplification
|   |   |-- inner_product.py
|   |-- gp_regressor.py       # Genetic programming for regression
|   |-- poly_regressor.py     # Polynomial tensor regression
|   |-- rl_regressor.py       # Reinforcement learning based discovery
|   |-- drsr/                 # LLM-based symbolic regression
|   |   |-- llm.py
|   |   |-- agent.py
|   |   |-- evaluator.py
|   |   |-- ...
|   |-- modules/              # Task-based neural modules
|   |   |-- TNlinear.py
|   |   |-- TNconv.py
|   |   |-- ...
|   |-- mlpregressor.py       # Task-based network for regression
|   |-- mlpclassifier.py      # Task-based network for classification
|   |-- neurons.py            # Neuron expression parsing & parameterization
|   |-- activation_function.py
|   |-- loss_function.py
|   |-- optimizer.py
|   |-- preprocessing.py      # DataPreprocessor
|   |-- utils.py
|-- tests/                    # Unit tests
|   |-- test_regressor.py
|   |-- test_mlpregressor.py
|   |-- test_mlpclassifier.py
|   |-- ...
|-- benchmark/                # Evaluation example scripts
|   |-- regression/
|   |-- classification/
|-- examples/                 # Usage examples
    |-- example_regression.py
    |-- example_classification.py
    |-- example_drsr.py
    |-- example_rl_regression.py
    |-- data/
\end{python}

\subsection{Installation}

\texttt{TNLearn} is available on PyPI, and can be installed via \texttt{pip}. It requires Python 3.8 or later. The library is built on top of PyTorch, scikit-learn, and NumPy, all of which are automatically resolved during installation.

\begin{python}
pip install tnlearn
\end{python}

For reproducibility, a specific version can be installed as follows:

\begin{python}
pip install tnlearn==0.1.0
\end{python}

Alternatively, the latest development version can be installed directly from the GitHub repository:

\begin{python}
git clone https://github.com/NewT123-WM/tnlearn.git
cd tnlearn
pip install -e .
\end{python}

After installation, the package can be imported as follows:

\begin{python}
import tnlearn
from tnlearn import GPSymRegressor, LLMSymRegressor, RLSymRegressor, PolyTensorRegressor, MLPRegressor, MLPClassifier
\end{python}

For GPU support, users should install the PyTorch version compatible with their CUDA version before installing \texttt{TNLearn}.

The documentation, including the API reference and tutorial notebooks, is available at \url{https://tnlearn-documentation.readthedocs.io}.

\section{Architecture and Implementation Guide}

This appendix walks through the typical workflow of \texttt{TNLearn}, from preparing data to building and training a task-based network. Rather than serving as an exhaustive API reference, it aims to convey the design philosophy behind the library and how its components fit together.

\subsection{General Workflow}

The standard pipeline proceeds through a sequence of stages, as illustrated in Figure~\ref{fig:tnlearn_arch}. The workflow is intentionally modular: users may replace any component, including the neuron discovery method, the network architecture, the optimizer, or the loss function, without altering the overall structure.

Together, these stages form an end-to-end, task-oriented pipeline that carries users from neuron discovery to task-specific network construction and training. The process begins with loading and optionally preprocessing the dataset, for example, through the built-in \texttt{DataPreprocessor}. If a task-specific neuron is desired, one of the provided symbolic regressors can automatically distill a suitable neuron formula from the training data; alternatively, users may hand-craft a formula and bypass this step. The user then instantiates either \texttt{MLPRegressor} or \texttt{MLPClassifier} with the discovered or manually specified neuron expression, together with network hyperparameters such as hidden-layer sizes, activation functions, loss function, and optimizer. For greater architectural freedom, a customized network can also be assembled from \texttt{tnlearn.modules}. Once the network is constructed, it is trained by calling \texttt{fit} and then evaluated with \texttt{predict} and \texttt{score} for \texttt{MLPRegressor} and \texttt{MLPClassifier}. For networks built with \texttt{tnlearn.modules}, training follows a PyTorch-style paradigm: users construct the computational graph to define the forward pass and design the loss function, then hand-code the optimization loop.

\subsection{Core Usage Pattern}

The following minimal example illustrates the complete pipeline, beginning with the built-in symbolic regressor and then feeding the discovered formula into a task-based MLP regressor. The same structure applies when using \texttt{LLMSymRegressor}, \texttt{RLSymRegressor}, or \texttt{PolyTensorRegressor}, as well as when a manually defined formula is supplied as the neuron input.

\begin{python}
import numpy as np
from sklearn.model_selection import train_test_split
from tnlearn import GPSymRegressor, MLPRegressor

# 1. Prepare data 
X_train, X_test, y_train, y_test = train_test_split(X, y, test_size=0.2)

# 2. Automatically discover a neuron formula
neuron_finder = GPSymRegressor(pop_size=2000, max_generations=15)
neuron_finder.fit(X_train, y_train)
formula = neuron_finder.neuron   # e.g., "x**3 + x + 1.0"

# 3. Build task-based network
model = MLPRegressor(
    neurons=formula,
    layers_list=[64, 32],
    activation_funcs='relu',
    max_iter=200,
    batch_size=64,
    lr=0.001
)

# 4. Train and evaluate
model.fit(X_train, y_train)
r2 = model.score(X_test, y_test)
print(f"Test R^2 = {r2:.4f}")
\end{python}

\subsection{Integration with Scikit-learn Workflows}

\texttt{TNLearn} adheres to the \texttt{scikit-learn} estimator API, making it straightforward to integrate into the existing pipelines, grid searches, and cross-validation workflows. For example, a task-based classifier can be passed to \texttt{GridSearchCV} just as any standard scikit-learn estimator, enabling hyperparameter tuning with minimal additional code.

\subsection{Extending TNLearn}

\texttt{TNLearn} is designed to be extended. Users may customize neuron formulas by supplying any valid mathematical expression as the \texttt{neurons} argument, following the syntax documented online. For deeper modifications, \texttt{CustomNeuronLayer} can be subclassed to define a novel aggregation function, and new activation or loss functions can be registered by extending the dictionaries in \texttt{activation\_function.py} and \texttt{loss\_function.py}. The developer documentation provides step-by-step guidance for these customization paths.

\section{Vectorized Symbolic Regressors in TNLearn}
\label{techniques}

Task-based neuron design aims to discover a symbolic expression that serves as the aggregation function of a neuron. Given an input matrix \(\mathbf{X} \in \mathbb{R}^{d \times N}\) (\(d\) features, \(N\) samples, with each column corresponding to one sample) and a target matrix \(\mathbf{Y} \in \mathbb{R}^{d_{\text{out}} \times N}\) (with columns aligned to the samples), the goal is to find a symbolic expression \(\phi\) (or a combination of expressions) that minimizes the empirical risk:
\begin{equation}
    \min_{\phi} \; \left\| \Y - \phi(\X) \right\|_F^2,
\end{equation}
where \(\phi(\X)\) is learned by symbolic regression. This reflects the core idea of task-based neuron design: a neuron that already fits the raw data well provides a task-aligned inductive bias, so a network composed of such neurons is expected to fit better than one built from generic neurons. During learning, the symbolic expression may apply a homogeneous univariate operation element-wise to each feature, possibly followed by an aggregation such as summation, or it may include products of inner products to model relationships between features and produce cross-terms.

To automate this discovery, \texttt{TNLearn} provides four complementary regressors: a genetic programming engine (\texttt{GPSymRegressor}), an LLM-driven searcher (\texttt{LLMSymRegressor}), a reinforcement learning agent (\texttt{RLSymRegressor}), and a differentiable tensor-based regressor (\texttt{PolyTensorRegressor}). They rely on different search strategies, and can produce homogeneous symbolic formulas, thereby enabling the construction of task-specific neurons with built-in feature interactions.

\subsection{GPSymRegressor}

\texttt{GPSymRegressor} is a genetic programming engine \citep{cramer1985representation} that discovers a symbolic expression \(\phi\) in the form of trees. Each candidate solution is a parse tree. Its internal nodes are arithmetic operators \(\{+, -, \times\}\), unary negation, and the binary inner-product operator \(\langle \cdot, \cdot \rangle\), while its leaf nodes are either the variable \(\x\) or random real constants from a predefined range. Such a tree can be converted to a string and simplified using \texttt{SymPy}. The simplified string is then evaluated by substituting \(\x\) with every sample of the entire input matrix \(\X\), thereby enabling a vectorized computation.

Since \texttt{SymPy} cannot handle expressions containing inner products, we customize the operation logic and simplification rules. The inner-product operator \(\langle \cdot, \cdot \rangle\) takes two sub-expressions, each being a matrix of shape \(d \times N\) or a scalar. If both are matrices, the inner product is computed for each sample as the element-wise sum of products across the feature dimension, yielding a vector of length \(N\). Scalars, in contrast, are handled by broadcasting. This operator enables the generation of interaction terms, such as the aggregation function \((\w_1^T\x)(\w_2^T\x)\) for each sample vector.

Natural selection is the survival of the fittest. Inspired by biology, we draw lessons from the evolution of species. Since we aim for the formula that best fits the original data, we need to iterate through the genetic algorithm to gradually obtain the tree structure corresponding to the best-fitting formula. Therefore, we define the mean squared error (MSE) between the target and the output as the main component of fitness, and allow genes with lower fitness to be inherited. In addition, to avoid trivial solutions, we discard expressions with fewer than two occurrences of \(\x\). In genetic operations, parent selection can be based on ranking or tournament selection, both selecting the one with the lowest fitness from a specific sequence.

\begin{figure}[H]
    \centering
    \includegraphics[width=1\textwidth]{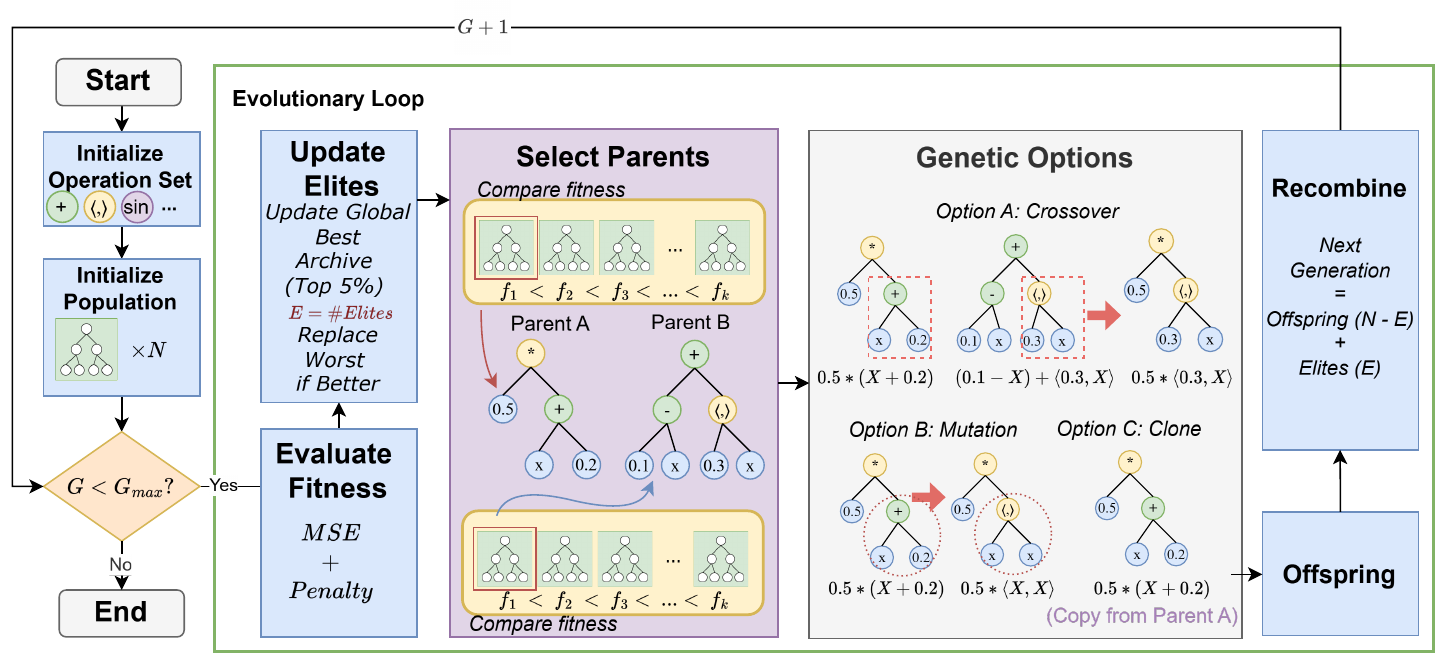}
    \vspace{-0.8cm}
    \caption{Genetic programming loop of GPSymRegressor, including evaluation, elite preservation, parent selection, genetic operations, and recombination.}
    \vspace{0.5cm}
    \label{fig:vsr_ops}
    \vspace{-0.8cm}
\end{figure}

Each generation has multiple genetic operations. Crossover exchanges random subtrees between two parents with probability \(p_{\text{xover}}\), mutation replaces a random subtree with a new one with probability \(0.9 - p_{\text{xover}}\), and reproduction copies a parent unchanged with probability \(0.1\). In addition, the program also sets a random immigration operation, which generates a completely different random tree. Subtree selection is biased toward deep nodes and away from the root to preserve large structures and enable local refinement. To maintain population quality, a small number of elite individuals are retained when entering the next generation. The overall workflow is illustrated in Figure~\ref{fig:vsr_ops}.

After the final generation, the expression \(\phi^*\) with the best (lowest) fitness is returned. The discovered formula provides an interpretable closed-form representation of the learned nonlinear transformation.

\subsection{LLMSymRegressor}

\begin{figure}[H]
    \centering
    \includegraphics[width=1\textwidth]{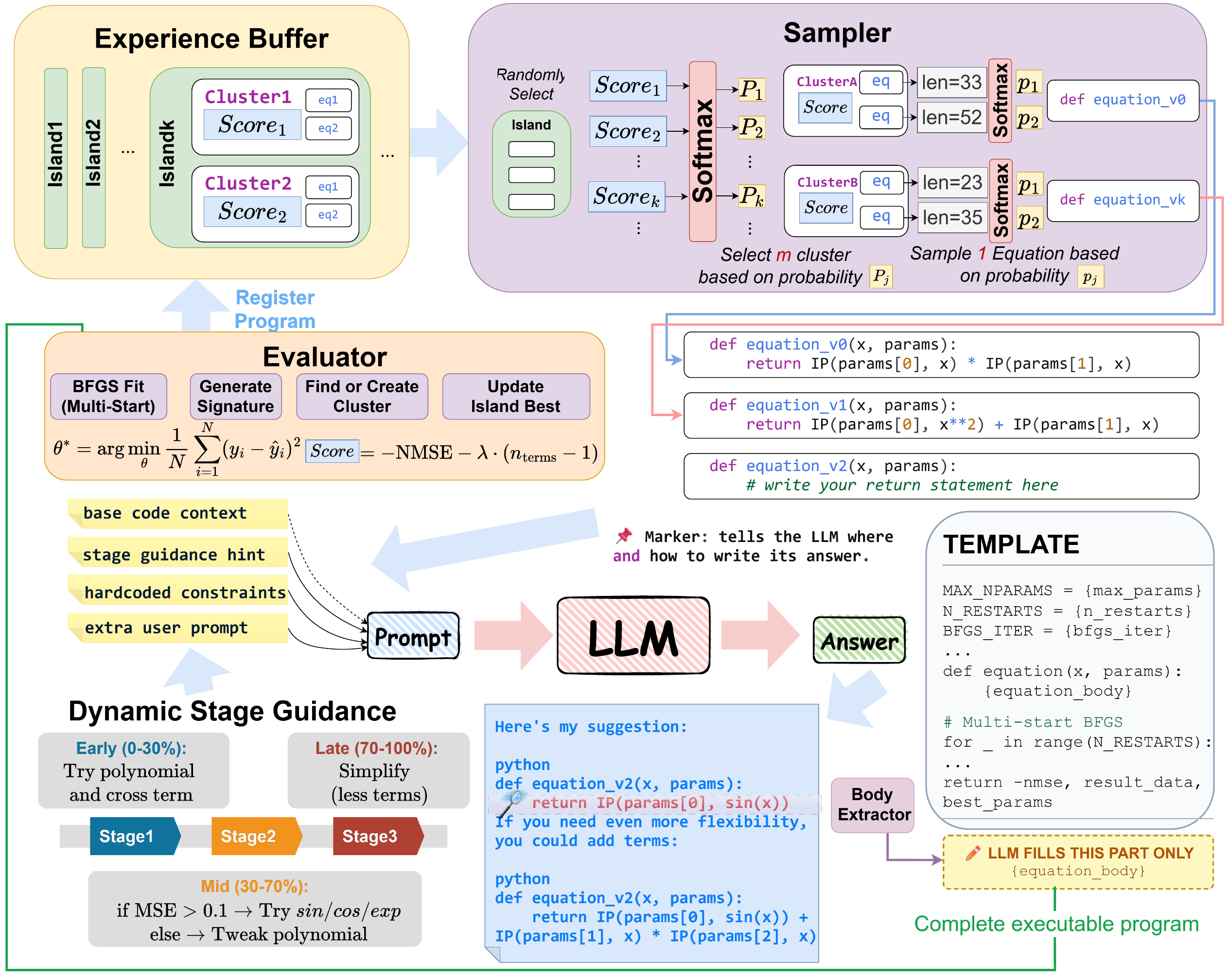}
    \vspace{-0.9cm}
    \caption{The workflow of \texttt{LLMSymRegressor}. The LLM generates candidate expressions from a prompt constructed by the experience buffer; each candidate is evaluated via BFGS and registered back into the buffer, forming a closed loop.}
\label{fig:llmsymregressor_ops}
\vspace{-0.5cm}
\end{figure}

\texttt{LLMSymRegressor} leverages a large language model to propose candidate symbolic expressions. In contrast to classical genetic programming, it does not search over tree structures. Instead, it exploits the LLM's prior knowledge and iterative prompting to generate new formulas. Each proposal is a Python function body of the fixed form ``return ...'', which depends on the variable \(x\) and a parameter array. This design allows the response to be seamlessly integrated with the evaluation program, unifying LLM-based proposal generation with formula evaluation and evolution into a single pipeline. With deliberate guidance and prompts that encourage diverse functional forms, the LLM can also generate expressive formulas. The workflow is illustrated in Figure~\ref{fig:llmsymregressor_ops}.

To guide the search, we maintain an experience buffer that records previously evaluated expressions and their scores. This buffer is organized as a multi-island model, where each island comprises multiple clusters of expressions with identical scores. These cached expressions facilitate prompt construction. The algorithm first randomly selects one of several islands (independent sub-populations) from the experience buffer, then samples clusters via a tempered softmax over their scores, and finally draws one expression from each selected cluster, where shorter expressions are preferred. This multi-level sampling provides the LLM with diverse, high-quality in-context demonstrations. Although performance evaluation and LLM querying are independent, after observing the existing high-quality expression structures, the LLM can directly propose potentially more suitable expressions. This capability stems from the LLM's internally accumulated understanding of underlying regularities. In addition, the prompt incorporates stage-wise hints according to the current search progress, encouraging polynomial terms in early stages and trigonometric or exponential functions later, while consistently favoring simplicity. This equips the regressor with the capability to discover complex formulas.

Once a candidate expression is generated, the evaluator assesses its quality by fitting its coefficients via BFGS with multiple restarts. The objective is to minimize the mean squared error on the training data. The score is computed as the negative normalized MSE on the training data to eliminate scale dependence. Then the best expression found during the search is retained, together with its optimized parameters.

The above steps form a closed loop. The algorithm iterates for a fixed number of sampling rounds. In each round, several new expressions are generated and evaluated. Meanwhile, the buffer periodically resets the weaker half of its islands and injects the best individual from the strongest island to maintain diversity. Finally, the algorithm outputs a symbolic expression in homogeneous aggregation form, with coefficients already optimized. The discovered formula can be directly used as a neuron in a task-based neural network, providing an interpretable nonlinear transformation.

\subsection{RLSymRegressor}

\begin{figure}[H]
    \centering
    \includegraphics[width=1.1\textwidth]{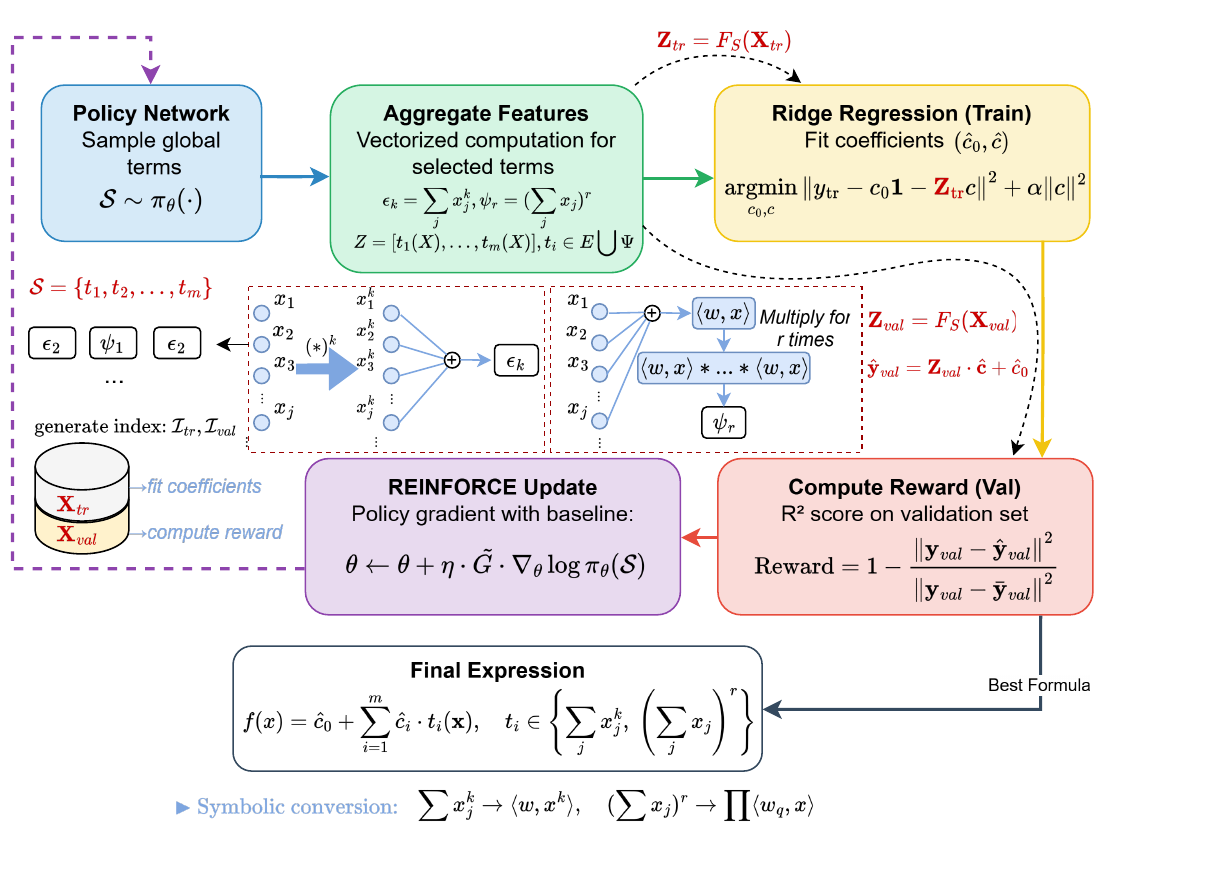}
    \vspace{-1.1cm}
    \caption{The workflow of \texttt{RLSymRegressor}. The policy network samples basis terms (\(\epsilon_k, \psi_r\)); selected terms are vectorized and Ridge-fitted; \(\mathrm{R}^2\) rewards in the validation set cause REINFORCE to update; the best expression is returned in inner-product form.}
    \vspace{-0.5cm}
    \label{fig:rl_ops}
\end{figure}

\texttt{RLSymRegressor} is a reinforcement learning agent that discovers a symbolic expression by selecting basis terms from a predefined library. Without considering coefficients, an aggregation function typically takes two common forms:
\begin{equation}
    \epsilon_k(\mathbf{x}) = \sum_{j} x_j^k \quad (k=1,\dots,K), \qquad
    \psi_r(\mathbf{x}) = \left(\sum_{j} x_j\right)^r \quad (r=2,\dots,R),
\end{equation}
where \(\mathbf{x}\) denotes the input vector, \(K\) denotes the user-specified maximum power, and \(R\) is the order of the interaction, respectively. Accordingly, we adopt these two families as the candidate library. The former provides power terms, and the latter provides interaction terms. Specifically, \(\psi_r\) captures multiplicative interactions among features, which are equivalent to inner-product-like combinations after expansion. At each episode, the agent samples \(m\) terms via a policy network. Duplicates may be retained because of the possible existence of multiple interaction patterns. Then, each term is assigned an independent Ridge coefficient. This reinforcement learning approach does not explicitly specify the coefficients but only selects a structure, letting the parameter optimizer optimize the formula as much as possible. This allows the network to focus on the structure rather than on freely adjustable parameters. The overall workflow is shown in Figure~\ref{fig:rl_ops}.

For a sampled set \(S = \{t_1, \dots, t_m\}\) with \(t_i \in \{\epsilon_k, \psi_r\}\), the regressor constructs feature vectors by applying each term directly to the input data \(\mathbf{X}\) using fully vectorized tensor operations along the feature dimension. This not only provides a fast aggregation operation but also eliminates explicit loops over individual features \(j\). The resulting columns are concatenated into a feature matrix \(\mathbf{Z}_{tr} = [t_1(\mathbf{X}_{tr}), \dots, t_m(\mathbf{X}_{tr})]\), which is then fitted using Ridge regression on the training set. The \(\mathrm{R}^2\) score in the validation set serves as the reward signal, and the policy is optimized via the REINFORCE \citep{williams1992simple} algorithm to maximize this reward. During training, the network learns a policy regarding how to select terms, aligning the selected structure with the goal of an optimal neuron aggregation function. The policy network is a multilayer perceptron that outputs a probability distribution over candidate terms. The REINFORCE update uses discounted returns with normalization to reduce variance, and the best expression over all episodes is retained.

The final output is a symbolic expression of the form:
\begin{equation}
f(\mathbf{x}) = c_0 + \sum_{i=1}^{m} c_i \, t_i(\mathbf{x}),    
\end{equation}
where \(c_0\) is the intercept, and \(c_i\) are the Ridge coefficients. It is then converted into an inner-product form, where each \(\epsilon_k\) becomes \(\langle \w, \mathbf{x}^k\rangle\), and each \(\psi_r\) becomes \(\prod_{q=1}^{r} \langle \w_q, \mathbf{x}\rangle\), with the \(\w\)'s serving as symbolic placeholders. The discovered expression is returned as a string, and can be directly used in neural networks due to its homogeneous aggregation form. The coefficients of the selected expression are later re-optimized by the network's parameter optimizer.

\subsection{PolyTensorRegressor}

\texttt{PolyTensorRegressor} provides a differentiable alternative to symbolic regression for discovering neuronal aggregation functions. By restricting the resultant expression to a tensor-decomposed form, it formulates neuronal discovery as a continuous optimization problem. Given an input matrix \(\mathbf{X}\in \mathbb{R}^{d \times N}\), the regressor represents a candidate aggregation function through three complementary streams. Specifically, the representation is given by
\begin{equation}
\begin{aligned}    \mathcal{A}_{\text{PolyTensor}}(\mathbf{X}) = \underbrace{\sum_{k=1}^{K} g_k^{(p)} \left( \mathbf{W}_k^\top \mathbf{X}^{\odot k} \right)}_{\text{polynomial stream}} + \underbrace{\sum_{m=2}^{M} g_m^{(i)} \sum_{r=1}^{R} \bigodot_{j=1}^{m} \left( \mathbf{A}_{r,j}^\top \mathbf{X} \right)}_{\text{interaction stream (CP decomposition)}} + \underbrace{g^{(s)} \mathbf{W}_s^\top \sin(\mathbf{X})}_{\text{periodic component}},
\end{aligned}
\end{equation}
where \(\mathbf{X}^{\odot k}\) denotes the element-wise \(k\)-th power of \(\mathbf{X}\), and \(\sin(\mathbf{X})\) is applied element-wise. In this expression, the matrices \(\mathbf{W}_k, \mathbf{A}_{r,j}, \mathbf{W}_s \in \mathbb{R}^{d \times d_{\text{out}}}\) are learnable projection matrices, and \(g_k^{(p)}, g_m^{(i)}, g^{(s)} \in [0,1]\) are differentiable soft gates that determine the extent to which each component is retained. The symbol \(\bigodot_{j=1}^{m} \left( \cdot \right)\) denotes the element-wise (Hadamard) product of the \(m\) matrices produced by the projections. Among the three streams, the polynomial stream captures independent nonlinear transformations of individual features via powers \(\mathbf{X}^{\odot k}\). The interaction stream models cross-feature dependencies of the order \(m \ge 2\) using a low-rank parameterization based on the CP decomposition~\citep{guo2011tensor}. Instead of explicitly constructing the full tensor and then decomposing it, we directly parameterize the interaction stream, which constrains the tensor rank and enables end-to-end learning of the factor matrices. This reduces the parameter complexity from \(\mathcal{O}(d^m)\) to \(\mathcal{O}(m R d)\) when \(d_{\text{out}}\) is treated as a constant, while preserving the ability to capture high-order feature interactions. The periodic component adds a sinusoidal term to handle cyclic patterns.

\begin{figure}[H]
\centering
\includegraphics[width=0.8\linewidth]{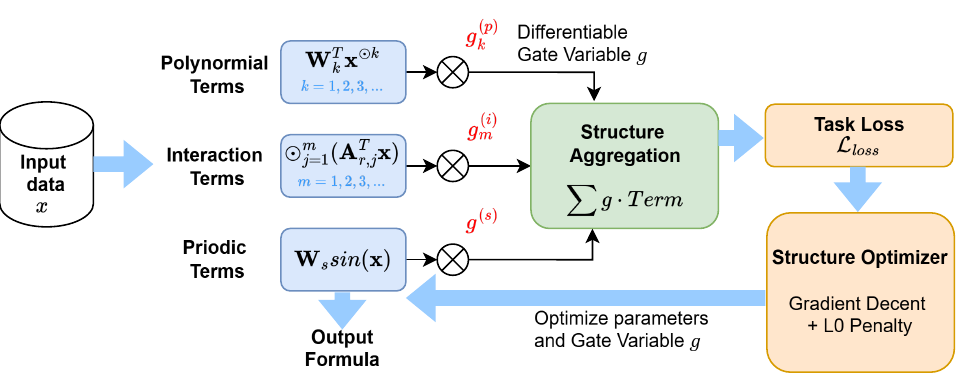}
\caption{The overall framework of the \texttt{PolyTensorRegressor}, showing the differentiable streams (polynomial, tensor-decomposed interactions, and periodic) and the gating mechanism for structural selection.}
\label{fig:poly-overview}
\end{figure}

The overall workflow is illustrated in Figure~\ref{fig:poly-overview}. To train the regressor, the gates \(g_k^{(p)}, g_m^{(i)}, g^{(s)}\) are optimized jointly with the projection matrices via gradient descent, using a sparsity-inducing regularizer such as an \(\ell_1\) penalty to prune redundant components~\citep{louizos2017learning}. After convergence, components are retained if their gates exceed a predefined threshold, and are pruned otherwise. This yields a compact formula form:
\begin{equation}
\phi_{\text{PolyTensor}}(\mathbf{x}) = \sum_{k} \mathbf{W}_k^\top \mathbf{x}^{\odot k} + \sum_{m} \sum_{r} \bigodot_{j=1}^{m} \left( \mathbf{A}_{r,j}^\top \mathbf{x} \right) + \mathbf{W}_s^\top \sin(\mathbf{x}),
\end{equation}
which can be directly parameterized as a trainable neuron layer. This differentiable search process improves the stability and reproducibility of discovered formulas compared to discrete genetic programming-based symbolic regression, especially for high-dimensional inputs or high-order interactions. The discovered neuron is returned in a form compatible with homogeneous aggregation, and can be embedded into neural network backbones without architectural redesign. 

\section{Discovered Neuron Formulas on Benchmark Datasets}

\renewcommand{\arraystretch}{1.3}
\setlength{\tabcolsep}{6pt}
\begin{table}[H]
\centering
\small
\resizebox{\linewidth}{!}{%
\begin{tabularx}{\dimexpr\linewidth+4em\relax}{@{}lcccX@{}}
\toprule
Dataset & Instances & Features & Task & \multicolumn{1}{c}{Discovered Formula} \\
\midrule
satimage & 6,430 & 37 & Classification &
$\begin{aligned}[t]
&-\big((\boldsymbol{0.0239}\odot\mathbf{x})^T\boldsymbol{0.8684}\big)+ (\boldsymbol{0.0239}\odot\mathbf{x})^T\mathbf{x}
\end{aligned}$ \\
phoneme & 5,404 & 6 & Classification &
$\begin{aligned}[t]
&0.0394\sum_i x_i^2 + 0.0243\sum_i x_i
\end{aligned}$ \\
adult & 48,842 & 15 & Classification &
$\begin{aligned}[t]
&-2\sum_i x_i + \mathbf{x}^T\boldsymbol{0.1406}
\end{aligned}$ \\
\midrule
pol & 15,000 & 49 & Regression &
$\begin{aligned}[t]
&-0.1301\sum_i x_i + 0.0139 - \boldsymbol{0.0039}^T\mathbf{x}
\end{aligned}$ \\
fried & 40,768 & 11 & Regression &
$\begin{aligned}[t]
&-1.7527\sum_i x_i^3 + 2.7958\sum_i x_i^2 \\
&-0.1638\sum_i x_i - 0.4164
\end{aligned}$ \\
bike & 17,379 & 12 & Regression &
$\begin{aligned}[t]
&(\mathbf{x}^T\boldsymbol{0.1966})(\boldsymbol{0.9731}^T\mathbf{x}^{\odot2}) - (\mathbf{x}^T\boldsymbol{0.1966})(0.9731\,\mathbf{x}^T\mathbf{x}) \\
&- (\mathbf{x}^T\boldsymbol{0.7017})(\boldsymbol{0.9731}^T\mathbf{x}^{\odot2}) + (\mathbf{x}^T\boldsymbol{0.7017})(0.9731\,\mathbf{x}^T\mathbf{x})
\end{aligned}$ \\
\bottomrule
\end{tabularx}}
\caption{Formulas discovered by \texttt{GPSymRegressor} on six benchmark datasets. Boldface numerals denote constant vectors; \(\mathbf{x}\) denotes the input vector; \(\odot\) denotes the element-wise product, and \(\mathbf{x}^{\odot k}\) the element-wise \(k\)-th power of \(\mathbf{x}\).}

\label{tab:discovered_formulas}
\end{table}


To evaluate the effectiveness of the task-based neuron framework implemented by \texttt{TNLearn}, we conduct experiments on a collection of datasets sourced from the OpenML repository \citep{casalicchio2019openml}. These datasets span a variety of application domains, sample sizes, and feature dimensionalities, providing a fair and comprehensive benchmark. For reproducibility, we use the OpenML ID of each dataset. The datasets used in this work include satimage (ID 182), phoneme (ID 1489), adult (ID 1590), pol (ID 201), fried (ID 564), Bike Sharing Demand (ID 44063), and sarcos (ID 43873). For brevity, we abbreviate Bike Sharing Demand as ``bike''.

Given these datasets, the first step is to apply vectorized symbolic regression to discover a formula that captures the underlying pattern of the data. Table~\ref{tab:discovered_formulas} lists the formulas automatically obtained by \texttt{GPSymRegressor} on six of these datasets. This provides an intuitive example of the formulas that the regressor can discover. For each dataset, we report the number of instances, the number of features, the task type, and the discovered formula. To maintain computational efficiency, we subsample up to 1,000 training instances during the formula searching stage. The numeric constants shown in the table are not fixed. When constructing task-based neurons, they are replaced by learnable parameters, as described in Section~\ref{sec:task_neuron_construction}. The resulting formulas then serve as the \texttt{neurons} argument when building a task-based network with \texttt{MLPRegressor} or \texttt{MLPClassifier}.

\section{Experimental Results with TNLearn}
\subsection{Evaluation on Classification and Regression Tasks}
We evaluate the effectiveness of the \texttt{TNLearn} library on both regression and classification tasks. All task-based networks are constructed using the \texttt{MLPRegressor} or \texttt{MLPClassifier} modules with neuron formulas discovered by \texttt{TNLearn}'s regressors.

A comprehensive comparison between task-based neurons and standard linear neurons is conducted with the same network architectures on the datasets. The results are summarized in Table~\ref{tab:ln_tn_comparison}. Across the evaluated datasets and architectures, the task-based neuron built by \texttt{TNLearn} consistently surpasses the linear neuron in both regression and classification tasks. This demonstrates that the task-based inductive bias embedded in each neuron via \texttt{TNLearn} significantly enhances feature representation power, eliminating the need to rely on simply stacking a large number of neurons to achieve competitive performance, which is a fundamental limitation of traditional linear networks.

\renewcommand{\arraystretch}{1.3}
\setlength{\tabcolsep}{5pt}
\begin{table}[H]
\centering
\small
\begin{tabular}{lcccc}
\toprule
Dataset & Architecture & Metric & Task-based Neuron & Linear Neuron \\
\midrule
satimage & 8--3--6 & Accuracy & \textbf{0.8896} & 0.8546 \\
phoneme & 6--2--2 & Accuracy & \textbf{0.8529} & 0.8104 \\
adult & 8--2 & Accuracy & \textbf{0.8582} & 0.8546 \\
\midrule
pol & 6--1 & \(\mathrm{R}^2\) & \textbf{0.9215} & 0.8926 \\
fried & 6--2--1 & \(\mathrm{R}^2\) & \textbf{0.9551} & 0.9389 \\
bike & 8--1 & \(\mathrm{R}^2\) & \textbf{0.8838} & 0.6702 \\
\bottomrule
\end{tabular}
\caption{The performance comparison of linear vs. task-based neurons. The architecture ``$n_1-n_2-\cdots-n_k$'' denotes the number of neurons in each layer, with the last layer being the output layer. For classification tasks, the output size equals the number of classes, while for regression, it is 1.}
\label{tab:ln_tn_comparison}
\end{table}

\renewcommand{\arraystretch}{1.3}
\setlength{\tabcolsep}{6pt}
\begin{table}[H] 
\centering
\small
\begin{tabularx}{\linewidth}{@{}lccX@{}}
\toprule
Dataset & Arch. (Params) & Score & \multicolumn{1}{c}{Symbolic Expression} \\
\midrule
\multirow{2}{*}{satimage} & 5--6 (1560) & 0.8989 & 
$\begin{aligned}[t]
&(\mathbf{w}_1^\top\mathbf{x})(\mathbf{w}_2^T\mathbf{x})(\mathbf{w}_3^T\mathbf{x}) + \mathbf{w}_4^T\mathbf{x}^{\odot 2} + \mathbf{w}_5^T\mathbf{x}^{\odot 3} + (\mathbf{w}_6^T\mathbf{x})(\mathbf{w}_7^T\mathbf{x})
\end{aligned}$ \\
 & 10--6 (1736) & 0.8087 & 
$\begin{aligned}[t]
&w_1 \sum_i x_i^3 + w_2 \sum_i x_i - (\mathbf{w}_3\odot\mathbf{x})^T\mathbf{x} + (\mathbf{w}_4\odot\mathbf{x})^T(\mathbf{1}\cdot \mathbf{x}^T\mathbf{x})
\end{aligned}$ \\
\midrule
\multirow{2}{*}{phoneme} & 5--2 (235) & 0.8548 & 
$\begin{aligned}[t]
&(\mathbf{w}_1^T\mathbf{x})(\mathbf{w}_2^T\mathbf{x}) + \mathbf{w}_3^T\mathbf{x}^{\odot 2} + \mathbf{w}_4^T\mathbf{x} + \mathbf{w}_5^T\mathbf{x}
\end{aligned}$ \\
 & 10--4--2 (356) & 0.8363 & 
$\begin{aligned}[t]
&w_1 \sum_i x_i^2 + w_2 \sum_i x_i + \mathbf{w}_3^T\mathbf{x} + \mathbf{x}^T\mathbf{x}
\end{aligned}$ \\
\midrule
\multirow{2}{*}{adult} & 5--2 (460) & 0.8610 & 
$\begin{aligned}[t]
&\mathbf{w}_1^T\mathbf{x} + (\mathbf{w}_2^T\mathbf{x})(\mathbf{w}_3^T\mathbf{x}) + \mathbf{w}_4^T\mathbf{x}^{\odot 3} + \mathbf{w}_5^T\mathbf{x}^{\odot 2}
\end{aligned}$ \\
 & 10--2 (704) & 0.8561 & 
$\begin{aligned}[t]
&w_1 \sum_i x_i + w_2 - w_3(\mathbf{x}^T \mathbf{w}_4) - (\mathbf{w}_5\odot\mathbf{x})^T(\mathbf{1}\cdot \mathbf{x}^T\mathbf{x})
\end{aligned}$ \\
\midrule
\multirow{2}{*}{bike} & 6--1 (469) & 0.9079 & 
$\begin{aligned}[t]
&\mathbf{w}_1^T \mathbf{x}^{\odot 3} + \mathbf{w}_2^T \mathbf{x}^{\odot 2} + (\mathbf{w}_3^T\mathbf{x})(\mathbf{w}_4^T\mathbf{x})(\mathbf{w}_5^T\mathbf{x}) + \mathbf{w}_6^T\mathbf{x}
\end{aligned}$ \\
 & 8--3--1 (506) & 0.8709 & 
$\begin{aligned}[t]
&w_1 - \big( (\mathbf{w}_2^T \mathbf{x})(\mathbf{x}^T\mathbf{x}) + w_3 (\mathbf{w}_2^T \mathbf{x})(\mathbf{1}^T\mathbf{x}) \big) - \mathbf{x}^T \mathbf{w}_4
\end{aligned}$\\
\midrule
\multirow{2}{*}{fried} & 5--1 (416) & 0.9586 & 
$\begin{aligned}[t]
&\mathbf{w}_1^T\mathbf{x}^{\odot 3} + \mathbf{w}_2^T\mathbf{x} + (\mathbf{w}_3^T\mathbf{x})(\mathbf{w}_4^T\mathbf{x})(\mathbf{w}_5^T\mathbf{x}) + (\mathbf{w}_6^T\mathbf{x})(\mathbf{w}_7^T\mathbf{x})
\end{aligned}$ \\
 & 10--4--1 (483) & 0.1597 & 
$\begin{aligned}[t]
&w_1 \sum_i x_i^4 + w_2 \sum_i x_i^2 - w_3(\mathbf{x}^T\mathbf{x}) + w_4(\mathbf{x}^T\mathbf{x})
\end{aligned}$ \\
\midrule
\multirow{2}{*}{pol} & 5--1 (1746) & 0.9908 & 
$\begin{aligned}[t]
&\mathbf{w}_1^T\mathbf{x} + \mathbf{w}_2^T\mathbf{x}^{\odot 2} + (\mathbf{w}_3^T\mathbf{x})(\mathbf{w}_4^T\mathbf{x}) + (\mathbf{w}_5^T\mathbf{x})(\mathbf{w}_6^T\mathbf{x})(\mathbf{w}_7^T\mathbf{x})
\end{aligned}$ \\
 & 10--1 (1981) & 0.9774 & 
$\begin{aligned}[t]
&w_1 \sum_i x_i + w_2(\mathbf{x}^T\mathbf{x}) - w_3 \sum_i x_i - (\mathbf{w}_4^T\mathbf{x})(\mathbf{1}^T\mathbf{x})
\end{aligned}$ \\
\midrule
\multirow{2}{*}{sarcos} & 5--1 (581) & 0.9710 & 
$\begin{aligned}[t]
&\mathbf{w}_1^T\mathbf{x} + \mathbf{w}_2^T\mathbf{x}^{\odot 2} + \mathbf{w}_3^T\mathbf{x} + (\mathbf{w}_4^T\mathbf{x})(\mathbf{w}_5^T\mathbf{x})
\end{aligned}$ \\
 & 10--1 (701) & 0.9693 & 
$\begin{aligned}[t]
&w_1 \sum_i x_i + w_2 + \mathbf{x}^T\mathbf{x} - ((\mathbf{x}^{\odot 2})^T \mathbf{w}_3)(\mathbf{w}_4^T \mathbf{x}^{\odot 3})
\end{aligned}$ \\
\bottomrule
\end{tabularx}
\caption{Comparison of task-based neurons and random neurons. For each dataset, the first row uses formulas discovered by \texttt{RLSymRegressor}, and the second row uses randomly generated formulas. Classification datasets (satimage, phoneme, adult) use Accuracy; regression datasets (Bike Sharing Demand, fried, pol, sarcos) use \(\mathrm{R}^2\). Architecture ``$n_1{-}n_2{-}\cdots{-}n_k$'' denotes the number of neurons per layer, with the last layer being the output layer, and the number of learnable parameters is shown in parentheses. All numeric constants are replaced by learnable parameters $w_i$ or $\mathbf{w}_i$ (vectors bold).}
\label{tab:neuron_comparison}
\end{table}
Furthermore, to verify if the formulas discovered by symbolic regression truly capture the underlying data patterns rather than arbitrary mathematical coincidences, we compare networks built with discovered formulas using \texttt{RLSymRegressor} against networks built with randomly generated formulas. The random formulas are generated by a tree-based generator using the same operators as \texttt{GPSymRegressor}, then subjected to multiple mutations and crossovers without fitness-based parent selection, and finally simplified with \texttt{SymPy}. For a fair comparison, they are generated with a comparable complexity to the discovered formulas in terms of the number of terms and occurrences of \(\x\). As shown in Table~\ref{tab:neuron_comparison}, the task-based network constructed by \texttt{TNLearn} consistently outperforms the random network across nearly all datasets, achieving higher regression \(\mathrm{R}^2\) or higher classification accuracy, even with smaller architectures. This advantage is achieved with fewer parameters, showing that task-based neurons can represent features effectively without increasing model capacity. These results also validate the symbolic regression engine in \texttt{TNLearn} as an effective extractor of task-relevant inductive biases, whereas random formulas lack such targeted representational effectiveness.

Despite the remarkable success of quadratic neurons in many tasks, \texttt{TNLearn} still achieves superior performance. We compare the proposed task-based neurons with four representative quadratic neuron variants \citep{bu2021quadratic, fan2018new, goyal2020improved, xu2022quadralib} on the benchmark datasets. We construct all these networks using \texttt{TNLearn} by specifying the corresponding quadratic neuron expressions. This provides a convenient way to benchmark different neuron designs. The task-based neurons compared here are those listed in Table~\ref{tab:discovered_formulas}. 

As shown in Table~\ref{tab:task_quad_net_result}, task-based neurons achieve superior or competitive performance. Unlike hand-crafted quadratic neurons with fixed parametric forms, \texttt{GPSymRegressor} automatically adapts the neuronal form to the characteristics of each dataset, providing greater flexibility. This flexibility is reflected in the diverse formulas discovered across datasets (Table~\ref{tab:discovered_formulas}). Overall, these results demonstrate that task-based neurons outperform classical quadratic neurons, which also establishes \texttt{TNLearn} as a convenient platform for benchmarking different neuron designs.

\renewcommand{\arraystretch}{1.3}
\setlength{\tabcolsep}{5pt}
\begin{table}[H]
\centering
\small
\scalebox{0.85}{\begin{tabular}{lcccccc}
\toprule
Dataset & Architecture & Ours & \citet{bu2021quadratic} & \citet{fan2018new} & \citet{goyal2020improved} & \citet{xu2022quadralib} \\
\midrule
pol & 12--5--1 & 0.9950 & 0.9762 & 0.9874 & 0.8678 & 0.9929 \\
fried & 4--1 & 0.9562 & 0.1586 & 0.1561 & 0.1589 & 0.9556 \\
bike & 8--3--1 & 0.9250 & 0.8910 & 0.9028 & 0.3880 & 0.9138 \\
satimage & 4--3--6 & 0.8756 & 0.7939 & 0.7426 & 0.6726 & 0.7729 \\
phoneme & 5--3--2 & 0.8548 & 0.7669 & 0.7928 & 0.7872 & 0.8131 \\
\bottomrule
\end{tabular}}
\caption{Comparison of task-based neurons and four quadratic neuron methods \citep{bu2021quadratic,fan2018new,goyal2020improved,xu2022quadralib}. Classification datasets (satimage, phoneme) use Accuracy; regression datasets use \(\mathrm{R}^2\). Architecture ``$n_1{-}n_2{-}\cdots{-}n_k$'' denotes the number of neurons in each layer, with the last layer being the output layer.}
\label{tab:task_quad_net_result}
\end{table}

\subsection{Evaluation on Images}

To validate the effectiveness of task-based neurons for convolutional networks in a controlled setting, we conduct experiments on the MNIST handwritten digit recognition dataset~\citep{lecun1998gradient}, which contains 60,000 training and 10,000 test grayscale images of size $28\times28$ across 10 classes. We construct a lightweight convolutional backbone consisting of convolutional layers, followed by global average pooling and a fully connected classifier. We compare two variants: the standard \texttt{Conv2d} in the PyTorch modules, which uses the conventional linear convolution $y = \mathbf{W} \ast \mathbf{x} + b$, and \texttt{TNConv2d} in the \texttt{TNLearn} modules, which replaces the standard convolution with a task-driven neuronal convolution. As a data type more complex than tabular data, images require more tricks for neuron discovery because their input dimensionality does not match that of the regressors. The aggregation function of this neuron is obtained through a two-stage discovery process. First, an autoencoder is used to reduce the dimensionality of image patches. Then, vectorized symbolic regression is applied to the resulting low-dimensional data. The autoencoder compresses the 2D spatially structured image into a compact low-dimensional latent vector while preserving essential task-relevant semantic information. It serves not only as a necessary bridge between image space and symbolic regression, but also provides dimensionality reduction, which reduces the computational cost of neuron discovery and improves convergence. 

Here, we discover a compact mathematical expression via \texttt{LLMSymRegressor}, which is subsequently implemented as the neuron formula. Both models share identical architecture, optimizer (Adam, learning rate $10^{-3}$), batch size (64), and training protocol (5 epochs). Each experiment is repeated 5 times, and the average results are reported.

\begin{table}[H]
\centering
\caption{Comparison of Standard \texttt{Conv2d} and \texttt{TNConv2d} on MNIST.}
\label{tab:mnist_comparison}
\renewcommand{\arraystretch}{1.1}
\begin{tabular}{cccccc}
\toprule
\multirow{2}{*}{Epoch} & \multicolumn{2}{c}{Standard \texttt{Conv2d}} & \multicolumn{2}{c}{\texttt{TNConv2d} (\texttt{TNLearn})} \\
\cmidrule(lr){2-3} \cmidrule(lr){4-5}
& Loss & Acc. (\%) & Loss & Acc. (\%) \\
\midrule
1 & 0.1614 & 98.49 & 0.1826 & \textbf{98.74} \\
2 & 0.0525 & 98.68 & 0.0568 & \textbf{99.02} \\
3 & 0.0394 & 99.05 & 0.0402 & 99.05 \\
4 & 0.0293 & 99.04 & 0.0308 & \textbf{99.18} \\
5 & 0.0252 & 99.05 & 0.0257 & \textbf{99.21} \\
\bottomrule
\end{tabular}
\end{table}

Table~\ref{tab:mnist_comparison} summarizes the training loss and test accuracy across all epochs. \texttt{TNConv2d} consistently achieves competitive performance while maintaining a comparable final training loss. In particular, \texttt{TNConv2d} exhibits faster early convergence.

\begin{figure}[H]
\centering
\includegraphics[width=0.7\linewidth]{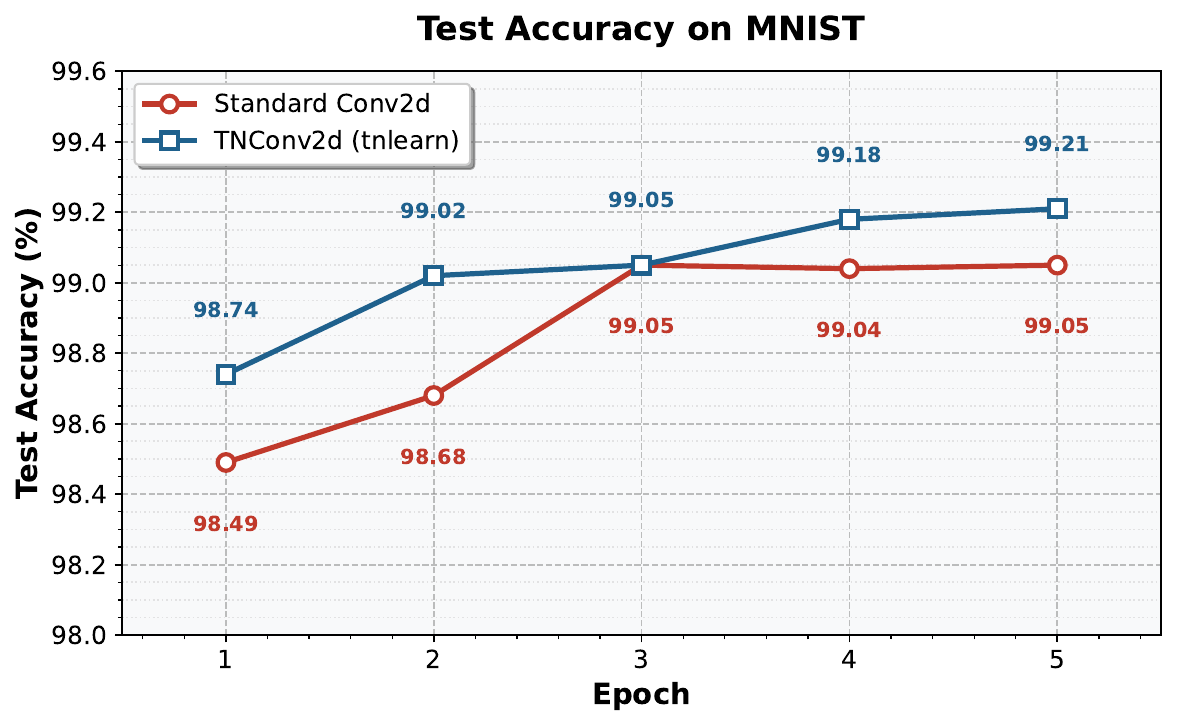}
\caption{The test accuracy curves for the standard \texttt{Conv2d} vs.~\texttt{TNConv2d} on the MNIST dataset over 5 epochs. }
\vspace{0cm}
\label{fig:mnist_comparison}
\vspace{-0.5cm}
\end{figure}

 The corresponding test accuracy curves are visualized in Figure~\ref{fig:mnist_comparison}. Task-specific symbolic neurons are able to capture more intricate pixel interactions than a purely linear kernel. The trend of the training loss shows that \texttt{TNConv2d} achieves a convergence speed comparable to that of the standard \texttt{Conv2d}, indicating that the increased complexity of the task-driven neuron does not compromise optimization efficiency. This indicates that the task-specific neuron design strikes a favorable balance between representational power and trainability.

\vskip 0.2in
\bibliography{sample}

\end{document}